\documentclass[10pt,journal,compsoc]{IEEEtran}

\ifCLASSOPTIONcompsoc
  \usepackage[nocompress]{cite}
\else
  \usepackage{cite}
\fi

\ifCLASSINFOpdf
\else
\fi

\usepackage{amsmath}
\usepackage{amssymb}
\usepackage{algorithm}
\usepackage{graphicx}
\usepackage{amsthm}
\usepackage{booktabs}
\usepackage{multirow}
\usepackage{color}
\usepackage{soul}
\usepackage{titlesec}
\usepackage[pdftex]{hyperref}

\titleformat{\paragraph}[runin]{\normalfont\normalsize\bfseries}{\theparagraph}{4em}{}[\hspace{4em}]

\usepackage[justification=centering]{caption}

\newcommand{\R}{\mathbb{R}}

\usepackage{algorithmic}

\usepackage{array}

\ifCLASSOPTIONcompsoc
  \usepackage[caption=false,font=footnotesize,labelfont=sf,textfont=sf]{subfig}
\else
  \usepackage[caption=false,font=footnotesize]{subfig}
\fi

\usepackage{url}

\begin{document}
%
\title{A Survey on Evaluation of Out-of-Distribution Generalization}

\author{
Han~Yu, Jiashuo~Liu, Xingxuan~Zhang, Jiayun~Wu, Peng~Cui$^\dagger$\thanks{$\dagger$ Corresponding Author}, \IEEEmembership{Senior Member,~IEEE}\thanks{ 
E-mail: yuh21@mails.tsinghua.edu.cn, liujiashuo77@gmail.com, xingxuanzhang@hotmail.com, jiayun.wu.work@gmail.com, cuip@tsinghua.edu.cn.
All authors are with the Department of Computer Science and Technology, Tsinghua University, Beijing, China. \protect
}
}

\markboth{}%
{Shell \MakeLowercase{\textit{et al.}}: Bare Demo of IEEEtran.cls for Computer Society Journals}

\IEEEtitleabstractindextext{
\begin{abstract}
Machine learning models, while progressively advanced, rely heavily on the IID assumption, which is often unfulfilled in practice due to inevitable distribution shifts. This renders them susceptible and untrustworthy for deployment in risk-sensitive applications. 
Such a significant problem has consequently spawned various branches of works dedicated to developing algorithms capable of Out-of-Distribution (OOD) generalization. 
Despite these efforts, much less attention has been paid to the evaluation of OOD generalization, which is also a complex and fundamental problem. Its goal is not only to assess whether a model's OOD generalization capability is strong or not, but also to evaluate where a model generalizes well or poorly. This entails characterizing the types of distribution shifts that a model can effectively address, and identifying the safe and risky input regions given a model. 
This paper serves as the first effort to conduct a comprehensive review of OOD evaluation. We categorize existing research into three paradigms: OOD performance testing, OOD performance prediction, and OOD intrinsic property characterization, according to the availability of test data. Additionally, we briefly discuss OOD evaluation in the context of pretrained models. In closing, we propose several promising directions for future research in OOD evaluation.

\end{abstract}

}

\maketitle

\IEEEdisplaynontitleabstractindextext

\IEEEpeerreviewmaketitle

\section{Introduction}
\label{sec:intro}

Over the past decade, significant advancements have transpired in the field of machine learning. Driven by the exponential growth of data and computational resources, neural networks have achieved marvelous performance across a wide spectrum of applications, including but not limited to computer vision \cite{he2016deep}, natural language processing \cite{devlin2018bert}, and recommender systems \cite{he2017neural}. 
Despite remarkable progress, it is imperative to acknowledge that existing machine learning algorithms and models still grapple with several persistent challenges that greatly undermine their reliability and trustworthiness. 
The challenges encompass issues such as privacy leakage \cite{shokri2015privacy}, weak interpretability of black box models \cite{zhang2018visual}, vulnerability under adversarial attacks \cite{akhtar2018threat}, and a notable decline in generalization performance when confronted with distribution shifts \cite{shen2021towards}. 
These aforementioned challenges pose a substantial impediment to the widespread adoption of current machine learning algorithms, particularly in fields of high risk sensitivity. 
Typical examples include law \cite{berk2021fairness}, where fairness and equity are paramount, medical care \cite{kukar2003transductive}, where patient well-being is at stake, and autonomous driving \cite{huval2015empirical}, where human lives depend on the performance of these algorithms. 

Among the challenges, one that poses a substantial obstacle is the problem of generalization against distribution shifts, often referred to as Out-of-Distribution (OOD) generalization. 
This is most prevalent because current algorithms heavily rely on the IID assumption, i.e. test data and training data should be independent and identically distributed, but distribution shifts are almost everywhere. 
In real applications, we can hardly guarantee that the test data encountered by deployed models will conform to the same distribution as training data. 
For example, models of recommender systems are trained on user data collected from the United States, but are tasked with extending to users in different countries, where the distribution of user preferences may vary largely. 
Similarly, visual recognition models trained on a dataset primarily comprising real photographs face the demanding task of recognizing images of diverse styles, such as art paintings, which represent a substantial distribution shift from their training data \cite{li2017deeper}. 
Additionally, demographic groups are usually imbalanced in training data, particularly with respect to gender or race. In such cases, the generalization performance of models can easily drop when minority groups in the training data dominate the test data \cite{ding2021retiring}. 
All these instances of distribution shifts bring about performance degradation of machine learning algorithms. 
Although domain adaptation techniques \cite{ben2010theory,long2015learning,long2017deep} have been developed to solve a similar problem with the help of test data since much earlier, in wild environments, distribution shifts are ubiquitous and unknown, where we are unlikely to have access to test data a priori.

One straightforward way involves developing algorithms that enhance the OOD generalization ability of models when test data is totally unknown, as highlighted by Shen et al. \cite{shen2021towards}. 
In recent years, several branches of works have been dedicated to this objective. 
Domain generalization (DG) \cite{wang2022generalizing,zhou2022domain,gulrajani2020search} takes advantage of multiple training domains to enable models to generalize to previously unseen test domains, mainly in the area of computer vision. 
Distributionally Robust Optimization (DRO) and its variants \cite{namkoong2016stochastic, sinha2018certifying, duchi2021learning} try to address the worst-case distribution. 
Invariant learning \cite{arjovsky2019invariant, liu2021heterogeneous, liu2021kernelized} seeks to capture the underlying heterogeneity and invariance present in the training data. 
Stable learning \cite{shen2020stable,kuang2020stable,yu2023stable} methods borrow ideas from causal inference to decorrelate variables through sample reweighting. 
These diverse branches collectively contribute to the overarching goal of improving OOD generalization, each offering unique insights and advancements that cannot be ignored. 

An alternative avenue for rendering machine learning models applicable in high-stakes areas is \textbf{evaluation}, i.e. evaluating their generalization ability under possible OOD scenarios. 
In contrast to the rapid proliferation of OOD generalization algorithms, much less attention has been paid to the aspect of evaluation. Evaluation is imperative across various areas of machine learning. Appropriate evaluation protocols and methods possess the potential to catalyze advancements in a field, just like ImageNet \cite{deng2009imagenet} did in the realm of computer vision. 
In the context of OOD generalization, evaluation plays an even more fundamental role. 
On one hand, OOD evaluation appears generally more intricate compared with its ID (In-Distribution) counterpart. For instance, given a dataset of a single task like image recognition, a natural approach entails randomly splitting it into a training set and a test set, with test accuracy serving as the evaluation metric for ID generalization. But when confronted with the same dataset, how can we split the dataset to generate the desired distribution shift and characterize the shift? Is this distribution shift solvable \cite{zhang2023nico++}? It becomes a much more complicated procedure within the context of OOD. 
On the other hand, it is worth noting that none of the current OOD generalization algorithms have achieved universal and tremendous improvement across various OOD settings and blasted the OOD community like ResNet \cite{he2016deep} did to the community of computer vision, or Transformer \cite{vaswani2017attention} did to the community of natural language processing. 
Indeed, it is quite difficult to develop a model that consistently outperforms others by a large margin in terms of OOD generalization, given that there are diverse types of distribution shifts to be addressed \cite{ye2022ood}, and one can hardly address them once and for all. 
Under this circumstance, it is more practical and useful to evaluate \textit{where} a model excels or falters. Specifically, our objective shifts towards identifying what types of distribution shifts there exist \cite{cai2023diagnosing} and assessing the model's competence in handling them, and we seek to pinpoint safe and risky input regions in which the model demonstrates exceptional or terrible performance \cite{liu2023need}. In this way, we can fully take advantage of existing trained models, which cannot generalize arbitrarily but are applicable to certain scenarios. 

Besides, OOD evaluation offers additional benefits when compared with directly developing a model tailored for OOD generalization. 
Nowadays, training a deep model from scratch, or even merely fine-tuning an existing one, could be time-consuming and expensive \cite{zhao2023survey}. In data-scarce scenarios like rare diseases \cite{lee2022deep}, there could be not enough data for training at all. In such cases, despite being unable to further improve it, we can decide whether and where to use this model with the help of proper OOD evaluation methods, or pick the suitable one from a model pool. 
Moreover, OOD evaluation is more flexible as well. The designed evaluation metric does not have to be optimizable, and it can be seamlessly incorporated into the process of model selection, which represents a crucial but seldom investigated aspect of OOD generalization \cite{gulrajani2020search,yu2023rethinking}. It can also be combined with non-algorithmic operations like adding extra data or features \cite{liu2023need}.  

Therefore, we consider OOD evaluation as a fundamental direction for OOD generalization. The goal is not only to determine whether a model possesses a good OOD generalization ability, but also to identify where it can perform well, including the types and degrees of distribution shifts, along with the corresponding input regions where the model maintains its competence. 
In this paper, we provide a systematic review of current OOD evaluation protocols, metrics, and methods, covering the multifaceted goals of OOD evaluation. To the best of our knowledge, we are the first to take a comprehensive view of this area. Previous surveys mainly concentrated on either the OOD generalization itself \cite{shen2021towards}, or on the general evaluation of machine learning models \cite{zhao2023survey}. Another survey also reviews the evaluation of OOD models but its scope is confined to the realm of NLP only \cite{li2023survey}. 

The rest of this paper is organized as follows. 
In Section \ref{sec:problem}, we introduce the problem setting and the categorization of OOD evaluation paradigms. 
In Section \ref{sec:testing}, \ref{sec:prediction}, \ref{sec:measurement}, we describe each category of OOD evaluation paradigms, mainly according to the dependence on OOD test data. 
In Section \ref{sec:pretrain}, we discuss OOD evaluation within the scope of pretrained models, including Large Language Models (LLMs). 
Finally, in Section \ref{sec:conclusion}, we conclude this paper and raise some directions that merit further exploration in future research endeavors. 

\section{Problem setting and categorization}
\label{sec:problem}

\begin{figure*}[ht] 
\centering
\includegraphics[width=1.0\textwidth]{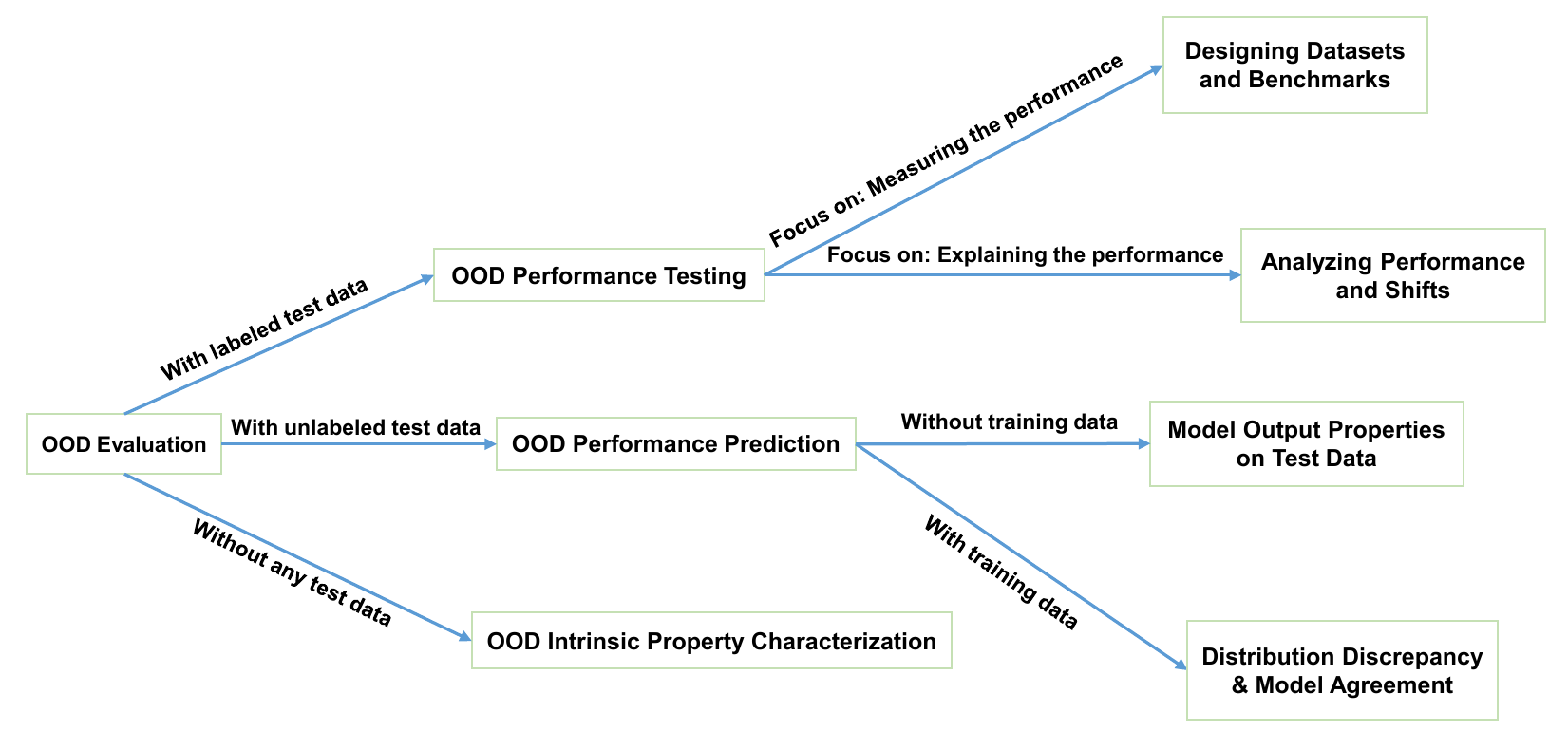} 
\caption{An overview of OOD evaluation. }
\label{fig:framework}
\end{figure*}

In this section, we first introduce the definition of the OOD generalization problem, and then we categorize existing literature on OOD evaluation into three distinct parts based on their dependence upon OOD test data. 

\subsection{Problem Definition of OOD Generalization}

Following Shen et al. \cite{shen2021towards}, we denote the input space as $\mathcal{X}$ and the output space as $\mathcal{Y}$. We define the parametric mapping function as $f_{\theta}: \mathcal{X}\mapsto \mathcal{Y}$, which represents the machine learning model. There is also a loss function describing the dissimilarity between the predicted output and the ground-truth output, denoted as $\ell:\mathcal{Y}\times \mathcal{Y}\mapsto \mathbb{R}$. As usually assumed, we consider test data originating from a probability distribution $P_{te}(X,Y)$, where both $X\in\mathcal{X}$ and $Y\in\mathcal{Y}$ are random variables. In our paper, we focus on the classical supervised learning problem. Consequently, we assume that training data is sampled independently and identically from a probability distribution $P_{tr}(X,Y)$. The primary goal is to find a model $f_{\theta}^*$ that exhibits optimal performance in the test distribution:

\begin{equation}
    f_{\theta}^*=\arg\min_{f_{\theta}} \mathop{\mathbb{E}}\limits_{X,Y\sim P_{te}} [\ell(f_{\theta}(X),Y)]
\end{equation}

When $P_{tr}(X,Y)=P_{te}(X,Y)$, the IID assumption holds, it becomes a standard learning problem. With the IID assumption, Empirical Risk Minimization (ERM) \cite{vapnik1991principles}, which directly minimizes the average loss of the training samples, can achieve a good performance. Conversely, when $P_{tr}(X,Y)\neq P_{te}(X,Y)$, a distribution shift occurs, transforming it into an OOD generalization problem. In such instances, ERM may fail severely in the face of distribution shifts \cite{torralba2011unbiased,castro2020causality}. 
It is crucial to emphasize that assumptions and restrictions regarding the types and degrees of distribution shifts need to be made. On one hand, it is obviously impossible for any model to generalize when test distribution bears no resemblance at all to training distribution. On the other hand, even for reasonable and solvable distribution shifts, it is hard for a model to address all of them simultaneously. 

\subsection{Categorization of OOD Evaluation}
It is difficult to define OOD evaluation mathematically since the evaluation is complex and can be carried out in various forms. It could be a prescribed protocol, a quantitative metric, or a predictive algorithm. Despite this, still we make an organized categorization of current works related to OOD evaluation based on the requirement for OOD test data: OOD performance testing, OOD performance prediction, and OOD intrinsic property characterization. Each of these paradigms represents a unique facet of OOD evaluation. The overall framework is shown in Figure \ref{fig:framework}. 

\paragraph{OOD Performance Testing}
The OOD performance testing paradigm involves the evaluation of models when labeled test data is available. While the operation of \textit{testing} a model on an OOD dataset seems rather straightforward, the intricacies lie in the evaluation protocol and the underlying design principles, e.g. how to generate distribution shifts or divide the dataset into distinct environments for various data types and tasks, and how to mitigate unfair comparisons stemming from oracle model selection as discussed by Gulrajani et al. \cite{gulrajani2020search}. 
Moreover, not only do we want to test how well models perform, but also we seek to comprehend the reasons behind a model's performance when confronted with the given distribution shift. This deeper understanding allows us to provide insights that can inform future algorithmic developments or non-algorithmic operations, such as data collection efforts targeting specific groups or attributes with the aim of enhancing OOD generalization \cite{liu2023need}. 

\paragraph{OOD Performance Prediction}

In practice, obtaining labeled OOD test data is often a formidable challenge. It is a common case that only unlabeled test data is available. Under this circumstance, we resort to OOD performance prediction as a viable strategy to select the best model or determine the usability of a given model with the help of unlabeled test data. 
Within this paradigm, researchers employ various strategies. Some rely solely on properties of model output on the given test data, e.g. model confidence \cite{garg2021leveraging}, without access to training data. 
Others adopt different approaches by harnessing information from the training data. They achieve this by characterizing the distribution discrepancy between the unlabeled training data and test data in a certain space induced by the evaluated model \cite{deng2021labels}, or calculating model agreement \cite{baek2022agreement} as a means of predicting test performance.

\begin{figure*}[ht] 
\centering
\includegraphics[width=1.0\textwidth]{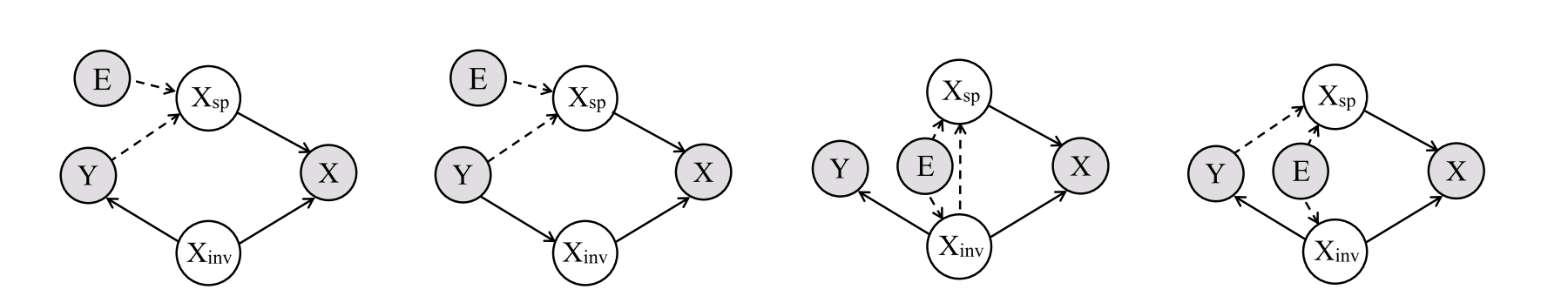} 
\caption{Common types of causal graphs when generating synthetic datasets. Image extracted from \cite{kaur2022modeling}. }
\label{fig:causal_graphs}
\end{figure*}

\paragraph{OOD Intrinsic Property Characterization}

When there is no test data available, OOD intrinsic property characterization tries to seek and delineate the inherent properties of models or training data that underpin OOD generalization. Current works have delved into properties such as distributional robustness \cite{li2021evaluating}, stability \cite{gupta2021s}, invariance \cite{ye2021towards}, and flatness \cite{zhang2023flatness}. 
These properties are closely connected with contemporary OOD generalization algorithms. For instance, distributional robustness is a direct derivative of DRO \cite{duchi2019distributionally}, invariance aligns with invariant learning \cite{arjovsky2019invariant}, and flatness intersects with flatness-aware optimizers \cite{foret2020sharpness}. 
It is worth noting that this pursuit is generally more challenging in the absence of test data, but it potentially probes deeper into the core essence of OOD generalization.

In the subsequent sections, we conduct a systematic review of these studies.

\section{OOD performance testing}
\label{sec:testing}

It is never a trivial problem to design the OOD performance testing procedure concerning a labeled dataset. For ID generalization tasks, given a labeled dataset, directly splitting it randomly into a training set and a test set will be reasonable enough. But for OOD generalization, many questions arise that require careful consideration. 
If there is no given dataset, how should we generate a labeled synthetic dataset from scratch to simulate a specific type of distribution shift? 
If there is a given dataset, how should we artificially generate a distribution shift or divide the dataset into multiple environments according to the existing distribution shift? 
After acquiring the dataset and environment partitions, how can we mitigate the potential possibility of test data information leakage resulting from pretraining \cite{yu2023rethinking} and oracle model selection \cite{gulrajani2020search}? 
Furthermore, once we have tested the model's performance on the labeled dataset, how should we analyze this performance effectively? 
Numerous issues need to be resolved in terms of OOD performance testing. 
In this section, we will introduce current OOD datasets or benchmarks along with their underlying design principles, and the existing methods to analyze the performance and distribution shifts. 

\subsection{Designing Datasets and Benchmarks}
In this section, we begin by categorically reviewing various types of OOD datasets respectively. Then we shift our focus to benchmarks composed of multiple datasets along with their protocols. 
\subsubsection{Synthetic Datasets}
Synthetic data is commonly exploited to assess an OOD method's effectiveness towards the targeted distribution shift. This involves the creation of an ideal environment through artificially designed data distributions.  

In general, a synthetic dataset is simulated according to a predesignated causal (bayesian) graph~\cite{neuberg2003causality} which describes the data generation process under the prospect of potential distribution shifts. 
As illustrated in Figure.~\ref{fig:causal_graphs}, distribution shifts are induced by the directed edge connecting the environment variable $E$ with the covariates $X$ and the target $Y$. 
The covariates $X$ can be decomposed into two distinct components: an invariant component denoted as $X_{inv}$ and a spurious component represented by $X_{sp}$, considering their respective associations with the target variable $Y$. Intuitively, a predictor relying on an invariant feature $X_{inv}$ achieves consistent performance across environments while predicting with a spurious feature $X_{sp}$ leads to a performance gap between different environments due to the presence of spurious correlations. Specifically, the correlation between $X_{sp}$ and either $Y$ or $X_{inv}$ could vary due to the intervention of $E$ in the causal graph. 

A causal graph synthesizes a family of data generation processes that adhere to a set of independence conditions based on the $d$-separation~\cite{neuberg2003causality} of the graph. A synthetic dataset is simulated through a unique data generation process, which is further formulated by Structural Equation Models (SEM)\cite{wright1921correlation}. For example, Invariant Risk Minimization (IRM)~\cite{arjovsky2019invariant} introduces a toy synthetic OOD dataset according to the fourth causal graph in Figure.~\ref{fig:causal_graphs}, whose SEM is given by:
\begin{align}
X_{inv} &\leftarrow \text{Gaussian}(0, E^2), \\
Y 	&\leftarrow X_{inv} + \text{Gaussian}(0, E^2),\\
X_{sp}  &\leftarrow Y+\text{Gaussian}(0, 1).
\end{align}
It can be verified that by simply regressing $Y$ on $(X_{inv},X_{sp})$ or $X_{sp}$, the estimated coefficients vary with values of $E$, demonstrating a spurious correlation. In contrast, the coefficients when regressing $Y$ on $X_{inv}$ are constantly $1$, because $X_{inv}$ is the cause of $Y$. Thus it becomes evident that $X_{inv}$ stands as the unique predictor of $Y$ that performs consistently across different environments.

\subsubsection{Visual Datasets}

Visual datasets are the most prevalent real-world datasets when it comes to OOD generalization, where image classification occupies the vast majority. One reason for the popularity of visual OOD datasets is that there naturally exist diverse types of distribution shifts with semantic meanings in computer vision. 
PACS \cite{li2017deeper}, OfficeHome \cite{venkateswara2017deep}, and DomainNet \cite{peng2019moment} generate shifts via the change of image style. 
WaterBirds \cite{wah2011caltech}, MetaShift \cite{liang2021metashift}, and ImageNetBG \cite{xiao2020noise} alter the backgrounds. 
Different locations or views of the taken photos are leveraged by TerraInc \cite{beery2018recognition} and iWildCam \cite{beery2021iwildcam}. 
Corruptions are used by ImageNet-C and ImageNet-P \cite{hendrycks2018benchmarking}. 
Some also directly collect data from different sources and treat each source as a distinct data distribution like VLCS \cite{khosla2012undoing} and CXRMultisite \cite{puli2021out}. 
Another common type of shift is generated through demographic attributes like race and gender, e.g. CelebA \cite{liu2015deep}, MIMIC-CXR \cite{johnson2019mimic}, and CheXpert \cite{irvin2019chexpert}. 
Note that most datasets cover only one kind of shift, rendering them relatively fixed and limited in testing OOD generalization performance. To promote greater flexibility in testing under various distribution shifts, it is encouraged to incorporate multiple attributes related to distribution shifts within a single dataset, like SVSF \cite{hendrycks2021many}, NICO++ \cite{zhang2023nico++}, and Spawrious \cite{lynch2023spawrious}. 

Apart from the shift type, the focus of visual OOD datasets differs for subpopulation shift \cite{yang2023change} and domain generalization \cite{gulrajani2020search}. Subpopulation shift datasets typically operate within a domain-imbalanced setting, deliberately creating situations where the majority group incurs an evident spurious correlation between the group label and the category label. These datasets primarily concentrate on evaluating the performance of the worst group, which usually corresponds to the minor group. As for domain generalization datasets, they are designed with relatively balanced data across domains, with a primary focus on the average performance across all test data instances. Both of these settings are common in real scenarios. 
There are also some datasets without their own training data, like ObjectNet \cite{barbu2019objectnet} and the variants of ImageNet \cite{hendrycks2021natural,hendrycks2021many,hendrycks2022scaling}. They are usually employed to test off-the-shelf models trained on ImageNet since the tested classes are a subset of ImageNet classes. 

Besides image classification, recently there have been various datasets tailored for other visual tasks. 
PovertyMap \cite{koh2021wilds} presents an image regression task in that its outcome label is the real-valued asset wealth index, whose environments are defined by urban or rural areas. Also introduced by Koh et al. \cite{koh2021wilds}, GlobalWheat aims at object detection to detect wheat heads in the given images in different countries, while COCO-O \cite{mao2023coco} targets a more general object detection with 80 object categories. SYNTHIA \cite{ros2016synthia} and GTA5-Cityscapes \cite{richter2016playing,cordts2016cityscapes} serve as datasets for OOD semantic segmentation. For more datasets concerning specific visual tasks, you may refer to the surveys on domain generalization \cite{wang2022generalizing,zhou2022domain}.

\subsubsection{Text Datasets}
Although less investigated than visual tasks, there are also various text datasets for OOD evaluation in NLP. Similar to visual datasets, a common design is to generate spurious correlations between attributes and outcome labels. For example, CivilComments \cite{koh2021wilds} poses the task of classifying whether the comment is toxic when there are correlations between the outcome and demographic attributes mentioned in the comment, such as gender and race. MultiNLI \cite{williams2018broad} takes advantage of the co-existence of negation words and the conclusion of contradiction to conduct natural language inference. 
Another type of design involves assembling datasets from diverse sources and subsequently dividing them into training and test sets. In Hendrycks et al. \cite{hendrycks2020pretrained}, the authors treat SST-2 \cite{socher2013recursive} and IMDb \cite{maas2011learning}, which contain movie reviews collected either by Stanford University or from IMDb website, as training and test datasets respectively. In Koh et al. \cite{koh2021wilds}, they partition the dataset Amazon according to the source of the comment, i.e. the reviewer. In the newly released benchmarks GLUE-X \cite{yang2022glue} and BOSS \cite{yuan2023revisiting}, distribution shifts also stem from the variation of collection sources. 

\subsubsection{Tabular Datasets}
Compared with the growth of visual and text datasets, tabular datasets are much less produced and remain underexplored. 
Temporal shifts and spatial shifts are commonly used. Shen et al. \cite{shen2020stable2} and Yu et al. \cite{yu2023stable} split the house price dataset \cite{house-prices-advanced-regression-techniques} according to the time period of the house built to generate distribution shifts. 
Liu et al. \cite{liu2023need} split the datasets Taxi \cite{nyc-taxi-trip-duration} and US Accident \cite{moosavi2019countrywide} according to the state or the year. 
In addition to spatial-temporal shifts, another natural shift stemming from demographic attributes are also adopted to introduce distribution shifts. 
In the UCI Adult dataset \cite{kohavi1996scaling}, a combination of race and sex can be used to formulate subgroups. 
Recently, the Bank Account Fraud (BAF) dataset \cite{jesus2022turning} is generated based on an anonymized dataset with the help of Generative Adversarial Networks (GAN) \cite{goodfellow2020generative} and differential privacy techniques and provides challenges concerning temporal shifts and data imbalance. This sheds light on the feasibility of creating larger-scale tabular datasets via data generation to facilitate more comprehensive evaluation.

\subsubsection{Benchmarks with Combined Datasets or Settings}
Apart from the aforementioned individual datasets, to further facilitate the development of OOD generalization methods, benchmarks combining these datasets with proper evaluation protocols are required. 
For most benchmarks, they directly adopt the average performance on the test environments as the evaluation metric. When it comes to subpopulation shift tasks \cite{yang2023change}, there is a particular emphasis on assessing the worst group performance. 

DomainBed \cite{gulrajani2020search} currently stands as the widely recognized evaluation protocol and benchmark for domain generalization, whose datasets are all tailored for image classification. It employs a leave-one-domain-out strategy, wherein each domain serves as the test domain while the remainder are utilized as training domains. 
Not only does it provide a codebase with strong flexibility enabling the implementation of new algorithms and put previous algorithms together for a fair comparison, but it also pioneers the investigation of a pivotal issue in OOD generalization, namely model selection. It shows the performance gain brought by potential information leakage when using test data for hyperparameter tuning, and designs an automatic hyperparameter random search pipeline that greatly helps mitigate this issue. It advocates model selection using validation data that shares the same distribution with the training one, and leaves the pursuit of a better OOD model selection method as an open problem. 

SubpopBench \cite{yang2023change} is a newly released benchmark for subpopulation shift, consisting of both visual and text datasets. 
Similar to DomainBed, it also integrates an automatic hyperparameter random search to alleviate the test information leakage. 
Previously, Idrissi et al. \cite{idrissi2022simple} pointed out that model selection utilizing group labels in validation data greatly contributes to performance improvements. SubpopBench also delves into the problem of model selection and discovers that simply using worst-class accuracy is quite effective even without oracle group labels, leaving ample room for designing a better metric without oracle information in the future. 

WILDS \cite{koh2021wilds} is another well-known benchmark featuring abundance in its tasks and data sources that cover a broad spectrum. 
It puts great efforts into the adaptation of previous large datasets to datasets conducive for OOD evaluation instead of directly employing existing datasets. 
The benchmark comprises 10 tasks of rich categories, including image classification, image regression, object detection, graph prediction, text classification, and code completion. The sources of images for the visual tasks range from camera-taken photos, satellite pictures, to medical imaging. 

Despite not being as large as the aforementioned three benchmarks, BREEDS \cite{santurkar2020breeds} and NICO++ \cite{zhang2023nico++} are visual benchmarks characterized by special design principles supporting rather flexible settings. BREEDS modifies the WordNet hierarchy based on visual characteristics to build a tree whose leaves correspond to ImageNet classes, and the internal nodes at varying depths in the tree correspond to different levels of superclasses. In this way, researchers can easily manipulate the granularity of the subgroup (subclasses) by adjusting the depth of the classes. Thus BREEDS generates 4 sub-tasks. 
For NICO++, each category is associated with 10 common domains and 10 unique domains. The common domains remain consistent across all categories, making them well-suited for standard domain generalization tasks. Meanwhile, the 10 unique domains of each category can be leveraged for more flexible settings. For example, researchers can choose some common domains as the majority domains and incorporate only a small amount of data from other domains to simulate the domain-imbalanced scenarios akin to subpopulation shift. In fact, a similar adaptation strategy has been adopted by SubpopBench \cite{yang2023change} to craft such a setting. 

Recently, GLUE-X \cite{yang2022glue} and BOSS \cite{yuan2023revisiting} have emerged as pioneering initiatives in benchmarking OOD generalization with NLP tasks. GLUE-X includes 8 tasks spanning across 15 datasets, and BOSS includes 5 tasks utilizing 20 datasets. Notably, there is limited overlap between the tasks and datasets utilized by these two benchmarks, rendering them highly complementary in nature. Compared with GLUE-X directly picking OOD datasets concerning each task and the corresponding ID dataset, BOSS conducts probing analyses using SimCSE scores \cite{gao2021simcse} and identifies pairs of datasets exhibiting the lowest semantic similarity, which are then employed as training and test pairs. Besides, GLUE-X focuses on benchmarking the OOD generalization performance of different pretrained models, and only tests three simple fine-tuning strategies. In contrast, more fine-tuning strategies are considered and tested in BOSS. Neither of them delves into the crucial issue of model selection issue, which is a prominent aspect of visual OOD benchmarks. 

In the realm of tabular data, although different kinds of natural shifts are frequently employed, current tabular datasets do not adequately capture specific distribution shift patterns (e.g, $Y|X$-shifts and $X$-shifts), undermining the practical significance of algorithm evaluations. 
Recently, to remedy this gap, Liu et al. \cite{liu2023need} have released WHYSHIFT. 
Drawing from open-sourced datasets such as Taxi \cite{nyc-taxi-trip-duration}, US Accident \cite{moosavi2019countrywide}, and the ACS dataset \cite{ding2021retiring}, the WHYSHIFT benchmark \cite{liu2023need} encompasses 22 settings that exhibit various \emph{defined} distribution shift patterns.
They conduct extensive experiments and find that different from visual and text data where covariate shifts on $P(X)$ dominate, concept shifts on $P(Y|X)$ are prevalent among tabular data. 
This conforms to our intuition since often there are missing variables that are not collected when establishing the tabular dataset. 
To investigate in-depth the shift pattern, they propose an algorithm to identify the risky regions, which could help to mitigate the effects of distribution shifts in a data-centric way. 
Different model selection strategies are also studied, and test information is found to be useful in improving OOD performance, sharing a similar conclusion to DomainBed \cite{gulrajani2020search} and SubpopBench \cite{yang2023change}.

\subsubsection{Discussion}

It is a delicate process to create OOD datasets or benchmarks, involving various ways of data collection. Some directly combine datasets from disparate sources \cite{khosla2012undoing,yuan2023revisiting}, some partition an existing dataset into different environments \cite{liu2015deep, santurkar2020breeds}, some leverage search engines for data acquisition \cite{recht2019imagenet,zhang2023nico++}, and some take new photos with required attributes \cite{barbu2019objectnet}. 
Although semantic perturbations are widely adopted for visual datasets to generate distribution shifts, their application to tabular data is notably challenging since it is difficult to ensure that the outcome variable remains constant when perturbing the covariates. 
Besides, we note that Spawrious \cite{lynch2023spawrious} is completely generated through stable diffusion models. It is worth investigating how data generation through SOTA generative models can help in creating OOD datasets. 
Moreover, despite the richness of shifts that existing datasets can accommodate, it is not clear whether such shifts described by predefined attributes or collection sources can simulate real scenarios well enough. For instance, there may exist more heterogeneous environment partitions that exhibit a larger and more challenging distribution shift than those defined by readily available environment labels \cite{liu2021heterogeneous}. 
Overall, much room is left for the design of OOD datasets and benchmarks.

\subsection{Analyzing Performance and Shifts}
After benchmarking the performance of OOD generalization algorithms against various distribution shifts, it is natural to take a further step to analyze the underlying reasons behind their performance and interpret these distribution shifts. This endeavor serves to foster a more profound comprehension of the factors that contribute to the success or failure of algorithms, thereby offering valuable insights for future development of OOD generalization algorithms. 

In line with this, Budhathoki et al. \cite{budhathoki2021did} incorporate causal graphs to investigate the cause of distribution shifts. 
They utilize Shapley values to attribute marginal distribution changes to each covariate. 
As demonstrated in a case study, this approach assists in identifying the driving factors behind differences, such as income prediction disparities between men and women.
Similarly, Zhang et al. \cite{zhang2022did} employ Shapley values to quantify the contribution of various distributions to the total performance change.
In addition to Shapley values, Kulinski and Inouye \cite{kulinski2023towards} frame the distribution shift as a transport problem and derive an interpretable relaxed optimal transport solution to account for the shifts.

Aside from directly analyzing dataset shifts, there have been efforts to dissect performance degradation in a more detailed manner. Cai et al. \cite{cai2023diagnosing} introduce the Distribution Shift Decomposition (DISDE) method to attribute performance decline to different types of distribution shifts, including $X$-shifts and $Y|X$-shifts. Leveraging shared distribution information, this study explicitly quantifies the performance drop caused by each kind of shift.
In response to performance degradation, to offer guidance for further improvements, Liu et al. \cite{liu2023need} suggest identifying high-risk regions or subpopulations that are prone to misclassification. Based on this identification, effective non-algorithmic interventions could be designed. Furthermore, they introduce the WHYSHIFT benchmark \cite{liu2023need} with specified shift patterns to evaluate algorithms designed for different types of distribution shifts.

While there has been growth in the analysis and interpretation of distribution shifts, applying these techniques to visual or text data remains challenging. For these types of data, concrete case analyses of misclassified examples are adopted at an early stage \cite{hendrycks2021natural}. Recent efforts have been made to formally describe and calculate such shifts. For instance, NICO++ \cite{zhang2023nico++} proposes to quantify the covariate shift and concept shift. SubpopBench \cite{yang2023change} classifies shifts into four distinct patterns and employs metrics such as mutual information and entropy for further calculation. Besides, OOD-Bench \cite{ye2022ood} decomposes shifts into correlation shift and diversity shift. Overall, it is imperative to devote more attention to bridging the gap between shift analysis and real-world data applications. 

\section{OOD performance prediction}
\label{sec:prediction}

Different from OOD performance testing where test data is fully accessible, this field represents another dimension of research aimed at predicting the performance on the provided unlabeled OOD test data. 
We can illustrate this with an example of medical image intelligent systems. These systems serve as auxiliary tools to help doctors with their diagnoses. Suppose the models deployed on these systems are trained on patient data from Beijing and Shanghai. Now we want to explore the possibility of deploying these models and systems in Guangzhou hospitals. 
Therefore, we need to predict the OOD performance of these models when presented with unlabeled image data collected from patients in the new scenario. 
Current works attempt to tackle this challenge from two perspectives. Some studies directly focus on the properties of model output when applied to the test data, without using training data. In contrast, other works employ distribution discrepancy or model agreement with the help of training data. 
The former is generally more convenient and flexible, yet the latter typically yields superior predictions since they take advantage of more information.

\subsection{Performance Prediction using Model Output Properties on Test Data}

Many works primarily center on harnessing the properties of model output on the test data, including model confidence, prediction dispersity, invariance under certain image transformations, etc. 

Drawn inspirations from research in OOD detection \cite{hendrycks2016baseline}, some methods propose metrics related to model confidence, a most commonly used property of model output when predicting OOD performance. 
These methods often operate under the intuitive assumption that a model's performance is superior in regions characterized by high confidence levels. 

Garg et al. \cite{garg2021leveraging} estimate the classification error via Average Threshold Confidence (ATC), the expected fraction of test samples for which the model's scores fall below a threshold $t$:
\begin{equation}
    {\rm ATC}_{P_{te}}(s):=\mathop{\mathbb{E}}\limits_{x\sim P_{te}}[\mathbb{I}[s(f_{\theta}(x))<t]]
\end{equation}
Here $f_{\theta}$ outputs the probability vector corresponding to each category, while $s$ is the score function, chosen as either the maximum confidence value in the vector or the negative entropy. The authors determine the threshold $t$ that enables ${\rm ATC}_{P_{tr}(s)}$ to be equal to the training error. They also compare with directly averaging model confidence and find that ATC outperforms it by a large margin. 

Ng et al. \cite{ng2022predicting} define the concept of manifold smoothness. 
\begin{equation}
    \mu(f_{\theta}, x):=\max_{j\in\mathcal{Y}} \frac{|\{x'\in N_{\mathcal{M}}(x):f_{\theta}(x')=j\}|}{|N_{\mathcal{M}}(x)|}
\end{equation}
where $N_{\mathcal{M}}$ refers to the manifold neighborhood of test point $x$. This smoothness metric further utilizes model confidence information in the neighborhood of the test point compared with ATC which only utilizes model confidence information of the test point itself. In their paper, they generate the neighborhood in a straightforward way of data augmentations. They empirically show that the smoothness metric outperforms ATC in terms of OOD performance prediction. 

Guillory et al. \cite{guillory2021predicting} endeavor to predict OOD performance using the first-order information of model confidence, i.e. Difference of Confidence (DOC). Concretely, they calculate the difference between confidence metrics on test data and training data, where the confidence metric can be maximum confidence or negative entropy. Experiments also demonstrate that DOC outperforms methods that only utilize zeroth-order information of model confidence. 

Different from model confidence, another two works respectively take a view at the prediction dispersity \cite{deng2023confidence} and model output invariance under certain transformations \cite{deng2021does}. 
Deng et al. \cite{deng2023confidence} emphasize the usefulness of dispersity, which measures how diverse and well-distributed model predictions are, in making up for the shortcomings of model confidence when predicting OOD performance. They use the nuclear norm to characterize both confidence and dispersity, and show that the nuclear norm correlates well with OOD performance. 
Deng et al. \cite{deng2021does} explore the relationship between invariance of prediction results under certain transformations and the OOD test accuracy. They experiment with the rotation and grayscale transformations, and uncover that both types of invariance show a strong correlation with test accuracy, suggesting their utility in predicting OOD performance. 

Besides, Liu et al. \cite{liu2023neuron} examine the neuron behavior when generating the model output on test data. They propose Neural Activation Coverage (NAC) to measure the activation level of neurons given an input sample. Their underlying assumption is that highly activated neurons are less likely to exhibit errors or anomalies, thus reducing the likelihood of encountering bugs.

Despite the demonstrated efficacy of these methods, as discussed by Garg et al. \cite{garg2021leveraging}, they often struggle to effectively address the subpopulation shift. This challenge is particularly pronounced in methods relying on model confidence. 
When a majority subpopulation in the training data exhibits a strong correlation with the outcome labels, machine learning models are prone to learning these spurious correlations. 
Consequently, these models may falter when confronted with test data where a minority subpopulation dominates \cite{sagawa2019distributionally,he2021towards}, even if they maintain a high level of confidence in their predictions for the failure cases.
A similar phenomenon "hallucination" is also observed in LLM \cite{zhao2023survey}, where models may exhibit unwarranted confidence in entirely erroneous or fabricated situations.

\subsection{Distribution Discrepancy based Performance Prediction}
\label{sec:disdis}

Given access to unlabeled test data, a natural thought is to borrow ideas from the field of Domain Adaptation (DA) \cite{ben2010theory} by considering the distribution discrepancy between training data and test data in a space induced by the evaluated model. 
Compared with methods reliant solely on model output properties on test data, these approaches take advantage of more information embedded in the training data, thus showing greater potential. 

In one of the earliest works targeting the OOD performance prediction, Deng et al. \cite{deng2021labels} employ the Frechet distance as a measure of distribution discrepancy:
\begin{equation}
    {\rm FD}(\mathcal{D}_{tr},\mathcal{D}_{te}):=\|\boldsymbol{\mu}_{tr}-\boldsymbol{\mu}_{te}\|+{\rm Tr}(\boldsymbol{\Sigma}_{tr}+\boldsymbol{\Sigma}_{te}-2(\boldsymbol{\Sigma}_{tr}\boldsymbol{\Sigma}_{te})^{\frac12}) 
\end{equation}
Here $\mathcal{D}_{tr}$ and $\mathcal{D}_{te}$ represent the covariate distributions of training and test data, while $\boldsymbol{\mu}$ and $\boldsymbol{\Sigma}$ denote the corresponding mean vector and the covariance matrix. This equation is applicable when the covariates are assumed to follow Gaussian distributions. Authors empirically demonstrate that this method outperforms baselines using a confidence threshold. 

Yu et al. \cite{yu2022predicting} depict the distribution discrepancy differently. Given a trained model $f_{\hat{\theta}}$ parameterized by $\boldsymbol{\hat{\theta}}$, first they assign pseudo labels to the OOD test data via the predictions of $f_{\hat{\theta}}$, then they train a new model $f_{\tilde{\theta}}$ on the pseudo-labeled test data which is initialized from $\boldsymbol{\theta_0}$. Finally they train another reference model parameterized by 
$\boldsymbol{\theta_{\rm ref}}$ using training data, also initialized from $\boldsymbol{\theta_0}$, and they calculate $\|\boldsymbol{\tilde{\theta}}-\boldsymbol{\theta_{\rm ref}}\|$ as the projection norm. 
Intuitively, the projection norm signifies the norm of the disparity between two sets of parameters that have been trained on training data and test data respectively, thereby assessing the distribution discrepancy to some extent. In fact, in the regime of overparameterized linear models, they theoretically prove that projection norm captures more comprehensive information regarding the covariate shift than the confidence score, playing a similar role as a better discrepancy metric does. 

Chuang et al. \cite{chuang2020estimating} employ a two-stage approach. They train a new domain-invariant classifier by aligning the representation of training and test data, and then finetune this classifier by maximizing the disagreement between the new classifier and the model to be evaluated while still preserving the domain-invariant property. 
A recent work \cite{lu2023characterizing} also predicts OOD performance from the angle of distribution discrepancy. Instead of the covariate shift, the authors focus on pseudo-label shift, i.e. the distance between the predicted label distribution and the ground truth test label distribution.

\subsection{Model Agreement based Performance Prediction}

Model agreement was applied in model selection with the help of unlabeled data in binary classification tasks approximately two decades ago \cite{madani2004co}. 
More recently, a surprising and interesting connection between model agreement and the generalization abilities has come to light in Nakkiran et al. \cite{nakkiran2020distributional} and Jiang et al. \cite{jiang2021assessing}, while the latter presents a stronger version: 
For classification tasks, when two models sharing the same network architecture are independently trained on identical datasets with identical sets of hyperparameters, but with variations in random initialization or training batch order, the agreement in their output predictions on In-Distribution (ID) test data approximately equals the test accuracy.
Jiang et al. \cite{jiang2021assessing} provide theoretical substantiation for this observation. They demonstrate that if the ensemble model is well-calibrated, the aforementioned equality holds, and they empirically confirm that the calibration condition can indeed be satisfied in practical scenarios.

While model agreement has traditionally been explored within the ID context, several works follow up by extending it to the OOD performance prediction. Baek et al. \cite{baek2022agreement} combine model agreement and the observation of accuracy-on-the-line \cite{miller2021accuracy}. They observe that OOD agreement strongly correlates with ID agreement, i.e. agreement-on-the-line, and the line even shares approximately the same slope and bias with that of OOD accuracy and ID accuracy. Therefore, by fitting a linear regression model between OOD and ID agreement, they can use it to predict OOD accuracy from ID accuracy. They empirically show that this prediction algorithm outperforms some other types of methods like ATC \cite{garg2021leveraging} and projection norm \cite{yu2022predicting}. 

Chen et al. \cite{chen2021detecting} take advantage of model agreement in an iterative framework. Given the model $f$ to be predicted, in each iteration, it learns an ensemble of $N$ auxiliary models $\{h_i\}_{i=1}^N$ that are trained on the original training set $D$ and the pseudo-labeled potentially misclassified test set $R$, and then pick the data points as $R$ on which the trained model $f$ and the ensemble model $h$ disagrees. The new $R$ is labeled as the output of $h$, i.e. the majority vote of the $N$ auxiliary models. This iterative approach not only enables the estimation of OOD accuracy, but can also output the potential misclassified data points. However, this method bears larger computational burdens due to the necessity of training multiple auxiliary models. 

Though powerful and effective, neither accuracy-on-the-line nor agreement-on-the-line hold under all kinds of OOD scenarios, and Baek et al. \cite{baek2022agreement} do not consider such agreement-based methods as universally applicable. 
In Kirsch et al. \cite{kirsch2022note}, authors question the satisfaction of the ensemble model's calibration when the disagreement increases, and they modify the theoretical analysis in a Bayesian probabilistic context. 
In Jiang et al. \cite{jiang2023joint}, authors propose a new theoretical framework of feature learning and apply it to explore the boundary of agreement-based methods. They re-partition the data by the proportion of blue intensity to change the ratio of the number of rare features to that of dominant features, which in their theory will lead to the inconsistency between model agreement and test accuracy.

\subsection{Discussion}
OOD performance prediction is an important paradigm for evaluating the OOD generalization ability without doubt, and much improvement has been made in existing works from various perspectives. 
However, there have been few works discussing the relationship and difference between various types of methods. Trivedi et al. \cite{trivedi2023closer} empirically show the advantages of agreement-based methods compared with confidence-based ones. Other types of methods have not been fully discussed. 
Also, few works have well characterized the application range, i.e. the type of distribution shift under which the proposed method can exhibit a good OOD prediction performance. Most works implicitly address the covariate-shift setting, while Chen et al. \cite{chen2022unsupervised} study the case where both label shift and sparse covariate shift exist. 
Besides, we note that current works focus on evaluating models of different network architectures with ERM training. It would be beneficial to explore the application of OOD performance prediction techniques to existing OOD generalization algorithms beyond the scope of ERM. 

\section{OOD intrinsic property characterization}
\label{sec:measurement}

In high-stakes applications like healthcare, understanding an algorithm or model's limitations in advance is of utmost importance. 
For instance, when deploying a model for autonomous driving, it is critical for ML researchers to anticipate its performance across different road conditions, weather conditions, traffic rules, etc.  
The challenge lies in the fact that typically, only historical data is available, with limited insight into prospective driving environments (the test data). 
This circumstance introduces an even more complex problem compared to the aforementioned OOD performance testing and prediction.
The pressing question thus becomes: how can we comprehend the \emph{\bf \textit{inherent}} characteristics of ML models in such a way that facilitates generalization under potential distribution shifts? We introduce from four perspectives: Distributional robustness, stability, invariance, and flatness. These perspectives connect closely to existing OOD generalization algorithms, like DRO, invariant learning, and flatness-aware optimizers, while their mutual connections are still under investigation. 

\subsection{Distributional Robustness}
\label{subsec:distributionalrobustness}
Emerging from the field of operations research, distributionally robust optimization (DRO) has attracted increased research attention in machine learning, which addresses the worst-case risk within a pre-defined uncertainty set.
The risk objective of DRO usually takes the form of
\begin{equation}
	\mathcal R(\theta):=\sup_{Q\in\mathcal P(P_{tr})}\mathbb E_{X,Y\sim Q}[\ell(f_\theta(X),Y)],
\end{equation}
where $\mathcal P(P_{tr})=\{Q:\text{Dist}(Q,P_{tr})\leq \rho\}$ denotes the uncertainty set surrounding the training distribution $P_{tr}$.
Note that no test information is used in the objective.
The distribution-wise distance metric, denoted as $\text{Dist}(\cdot,\cdot)$, offers multiple choices such as $f$-divergence, Wasserstein distance, or Maximum Mean Discrepancy (MMD) distance. 
Each choice corresponds to different DRO algorithms and theoretical guarantees. 
Intuitively, if the test distribution is encapsulated within the uncertainty set, it is plausible to assure the model's generalization performance.

In a parallel context, the principle of distributional robustness is leveraged to estimate the generalization potential of specific models. 
Li et al. \cite{li2021evaluating} evaluate model robustness against distribution shifts in a set of covariates $Z$ by examining the worst sub-population performance across all subpopulations of a given size. 
Considering a given model $f_\theta$ that derives from marginal DRO \cite{duchi2019distributionally}, they formulate their measure as follows:
\begin{equation}
	W_\alpha:=\sup_{Q_Z\in\mathcal Q_\alpha}\mathbb E_{Z\sim Q_Z}\bigg[\mathbb E[\ell(f_\theta(X),Y)|Z]\bigg],
\end{equation}
Here, $\mathcal Q_\alpha$ encompasses all the subpopulations with a proportion exceeding $\alpha$:
\begin{equation}
\begin{aligned}
	Q_\alpha:=\{Q_Z:P_Z&=aQ_Z+(1-a)Q_Z^{'} \text{ for some }a\geq \alpha \\&\text{and subpopulations } Q_Z^{'}\}.
\end{aligned}
\end{equation}
This proposed metric provides a means to assess robustness against \emph{covariate shifts} in any covariates of interest. The resulting worst-case subpopulation could thereby offer machine learning researchers guidance for further model improvement.
Building upon this, Subbaswamy and Saria \cite{subbaswamy2021evaluating} explore distribution shifts with more intricate covariate structures. 
They maintain the marginal distribution of immutable variables $Z\subset\{X,Y\}$ constant during covariate shifts, introducing distribution shifts solely to mutable variables $W\subset\{X,Y\}/ Z$. 
Compared with the measure in \cite{li2021evaluating}, an additional constraint is introduced to preserve $P(Z)$ unchanged. 
Addressing the conservatism problem associated with the uncertainty set, Thams et al. \cite{thams2022evaluating} incorporate a causal structure to parameterize the uncertainty set.
This approach extends beyond covariate shifts, also accommodating shifts in conditional distributions $Y|X$.
By means of a pre-defined causal structure, they are able to specifically consider distribution shifts pertaining to the specific edges of interest on the causal graph.

\subsection{Stability}
In conventional statistical analysis, a substantial body of literature is dedicated to sensitivity analysis. 
This field of study explores the stability of estimations under the influence of minor data perturbations.
Recently, there have been works \cite{gupta2021s,namkoong2022minimax} adapting this concept to assess model stability in the face of distribution shifts. 

Gupta and Rothenhäusler \cite{gupta2021s} propose a measure of instability that quantifies the minimal perturbation required in the data distribution to change the sign of the estimated parameters.
\begin{equation}
\begin{aligned}
		s(\theta,P_{tr}):=\sup_{P\in\mathcal P}\exp\bigg\{-D_{\text{KL}}(P\|P_{tr})\bigg\} \quad\text{s.t.}\ \theta(P)=0,
\end{aligned}
\end{equation}
where $D_{\text{KL}}(\cdot\|\cdot)$ denotes the KL-divergence.
Contrary to the distributional robustness notions in Section \ref{subsec:distributionalrobustness}, which quantify worst-case risks, this stability measure evaluates the magnitude of perturbations required to compromise an estimator.
Extending beyond parameter estimation, Namkoong et al. \cite{namkoong2022minimax} substitute the constraints with prediction error, investigating the extent of the distribution shift needed to degrade system performance to an unacceptable level.
\begin{equation}
	I_t(P_{tr}):=\inf_Q \bigg\{D_{\text{KL}}(Q\|P_{tr}): \mathbb E_Q[\ell]\geq t \bigg\},
\end{equation}
where $t>0$ is the threshold, and $\ell$ denotes the model prediction error.
This stability metric evaluates the tail performance with respect to the distribution $P_{tr}$.
Compared with distributional robustness, which requires the pre-specification of the uncertainty set's size, stability measures are more intuitive and interpretable. 
Machine learning engineers can more readily decide on a tolerable amount of prediction error based on their domain knowledge.

\subsection{Invariance}
In addition to distributional robustness and stability, there exists a diverse range of methods designed to determine whether the learned predictive relationship remains invariant across varying training environments or subpopulations.
These methods are natural extensions of invariant learning, where many algorithms have been proposed to learn the invariant relationship between covariates and labels.
Actually, the proposed penalties or regularizers in invariant learning algorithms \cite{arjovsky2019invariant, liu2021heterogeneous} could all be utilized to reflect whether a model captures the invariance or not.
In this section, we introduce some typical penalties in invariant learning, and one could refer to Shen et al. \cite{shen2021towards} for a detailed review.

Drawing inspiration from causal inference, Invariant Risk Minimization (IRM, \cite{arjovsky2019invariant}) extends Invariant Causal Prediction (ICP, \cite{ICP}) to more applicable settings. 
IRM \cite{arjovsky2019invariant} employs first-order information to quantify any violation of the invariance principle. 
The invariance penalty is approximated as follows:
\begin{equation}
	\sum_{e\in \mathcal E_{tr}}\bigg\|\nabla_{w|w=1.0}R^e(w\cdot \Phi)\bigg\|^2,
\end{equation}
Here, $\mathcal E_{tr}$ denotes the set of training environments, $\R^e$ represents the prediction risk in environment $e$, and $w, \Phi$ stand for the linear classifier and representation, respectively. 
When this measure equals 0, it indicates that the learned representation is optimal across the training environments.
Based on IRM, subsequent research (Section 4.1 in Shen et al. \cite{shen2021towards}) has proposed various forms of first- or second-order information to signify invariance.
Moreover, from a validation standpoint, Ye et al. \cite{ye2021towards} introduce a rigorous theoretical framework addressing the out-of-distribution (OOD) generalization problem. Through their theoretical analysis, they derive a model selection criterion to quantify the variation in learned representations.
Furthermore, drawing inspiration from multi-group calibration, Wald et al. \cite{wald2021calibration} establish a relationship between multi-group calibration and the invariance principle. 
As a result, they propose Calibration Loss Over Environments (CLOvE) for model selection.
When explicit environments cannot be predetermined, Liu et al. \cite{liu2021heterogeneous, liu2022measure} suggest initially exploring the latent heterogeneity within the data. 
Following this, the heterogeneous subpopulations can be incorporated with invariance penalties.

\subsection{Flatness}

The relationship between flatness and generalization has long been investigated \cite{keskar2016large, jia2020information, izmailov2018averaging, foret2020sharpness}. It is widely acknowledged that a flatter minimum leads to a better ID generalization ability. Moreover, recently there have been many works focusing on extending flatness to the area of OOD generalization, through either directly defining a flatness metric or an ensemble of network parameters to seek flatness. 

The most widely adopted flatness metric is proposed in Sharpness-Aware Minimization (SAM) \cite{foret2020sharpness}, defined as the maximal loss gap between the given point and the points in the neighborhood:
\begin{equation}
    R_{\rho}^{SAM}(\boldsymbol{\theta}):=\max _{\boldsymbol{\theta}'\in B(\boldsymbol{\theta},\rho)} \left(\hat{L}(\boldsymbol{\theta}')-\hat{L}(\boldsymbol{\theta})\right)
\end{equation}
where $\boldsymbol{\theta}$ is the given parameter point, $\rho$ denotes the radius of the neighborhood ball, $\hat{L}$ denotes the empirical loss. 
This metric only takes advantage of the zeroth-order information, i.e. the loss value itself. Thus Zhang et al. \cite{zhang2023gradient} propose Gradient-Aware Minimization (GAM), where a first-order flatness metric is defined as the maximal gradient norm in the neighborhood. 
\begin{equation}
    R^{GAM}_{\rho}(\boldsymbol{\theta}) := \rho \cdot \max_{\boldsymbol{\theta}' \in B(\boldsymbol{\theta}, \rho)} \left\|\nabla \hat{L}(\boldsymbol{\theta}')\right\|
\end{equation}
Compared with the metric of SAM, this GAM metric takes a step further by utilizing the first-order information of the loss gradient. Other metrics are also employed to describe flatness, like the maximal eigenvalue of the Hessian matrix $\lambda_{{\rm max}}(\nabla^2\hat{L}(\boldsymbol{\theta}^*))$ \cite{chaudhari2019entropy, lewkowycz2020large}. 
An extension of GAM is demonstrated to improve the ability of domain generalization \cite{zhang2023flatness}, revealing the potential of seeking flatness for OOD generalization. 
In the future, despite more complex flatness metrics, like metrics with a higher order, being hard to optimize, they can be designed for OOD evaluation, e.g. model selection.  

The other branch of works seeks flatness via an ensemble of network parameters. Stochastic Weight Averaging (SWA) \cite{izmailov2018averaging} averages the parameters every epoch or certain number of iterations after warming up. Authors empirically illustrate that such an averaging strategy leads to wider optima and stronger generalization. 
Based on this finding, Cha et al. \cite{cha2021swad} modify this strategy for DG tasks. They preserve the moving average of parameters across every iteration from the start of overfitting to the end of training. They also evaluate the flatness of trained models via the metric defined by SAM \cite{foret2020sharpness} and confirm the effectiveness of SWAD in improving flatness. 
Another DG method also employs moving average of network parameters \cite{arpit2022ensemble}. They demonstrate that the validation performance of moving averaged models is better and more stably correlated with OOD test performance than that of a single model. This indicates the averaged model's validation performance can be used for model selection. 
Recently many efforts have been put into designing ensemble strategies for DG in addition to averaging model parameters \cite{chu2022dna, wortsman2022model, rame2023model, rame2022diverse, li2022simple}. 

Despite more and more works focusing on this area, the relationship between flatness and OOD generalization abilities is not totally clear yet. For example, researchers find that in DG tasks, SAM does not outperform SWAD and even degrades the performance when combined with SWAD, although SAM is designed for achieving flatness more directly than SWAD \cite{cha2021swad, rame2022diverse}. Rame et al. \cite{rame2022diverse} point out the theoretical flaws of the relationship between flatness and OOD generalization error in SWAD, and provide a new analysis for model parameter averaging and OOD generalization error. Andriushchenko et al. \cite{andriushchenko2023modern} even discover a negative correlation between flatness and OOD generalization, contrary to our previous knowledge. 

\subsection{Discussion}
While we have explored the intrinsic properties from four distinct perspectives, they are intimately interconnected.
Distributional robustness is intricately linked to the notion of stability, where the objectives and constraints swap roles. The former quantifies the worst sub-population performance under predefined distribution perturbations, while the latter is characterized by the minimal perturbations necessary to compromise model performance.
Originating from the causal inference literature, invariance can be perceived as a milder form of causality. 
As highlighted in Meinshausen et al. \cite{meinshausen2018causality}, it represents a specific form of distributional robustness.
While the invariance property concentrates on model performance across varying environments, flatness is more attuned to the immediate vicinity, underscoring the model's local characteristics.

In general, these properties, though analyzed from different lenses, collectively form the bedrock of understanding models' resilience, adaptability, and reliability. Their intertwined nature underscores the necessity of a comprehensive approach to grasp the full spectrum of model behavior in diverse scenarios.

\section{Discussion on pretrained models}
\label{sec:pretrain}
Recently, pretrained models, especially Large Language Models (LLMs) \cite{brown2020language,bubeck2023sparks}, have achieved brilliant performance across a wide spectrum of tasks and garnered worldwide attention. 
However, results in the newly established OOD benchmark GLUE-X \cite{yang2022glue} show that even equipped with strong pretrained weights, models still fall far behind human-level performance in OOD generalization tasks. They also show that GPT-3.5, while surpassing GPT-3 by a large margin in standard NLP tasks, exhibits only marginal superiority in OOD tasks of GLUE-X. 
Moreover, GPT-3.5 even fails to outperform ELECTRA-large, a smaller pretrained model. A similar phenomenon is observed in another latest OOD benchmark BOSS \cite{yuan2023revisiting} that GPT-3.5 cannot always outperform smaller models. Besides, issues related to bias and fairness, which are closely related to OOD generalization, are still ubiquitous for LLM \cite{zhao2023survey,chang2023survey}. 
These results collectively underscore that the mere expansion of model parameters and training data, without concurrent development of OOD strategies, falls short of completely solving the OOD generalization problem.

However, as pretrained models continue to advance, their wide-spread and irreplaceable integration into various tasks naturally provokes a fundamental question concerning evaluating OOD generalization abilities: How should we define the distribution shift given a strong pretrained model that contains abundant information from its pretraining data? 
Indeed, it is possible that portions of the test data may have been encountered during the pretraining phase, or that the test distribution of input data closely aligns with that of the pretraining data. 
In such a case, models' performance drop on test data may not accurately reflect the distribution shift between test and training data, but rather the shift between test data and the combination of training and pretraining data. 
Previous works \cite{kumar2021fine,yu2023rethinking} have shown that simple linear-probing can outperform fine-tuning the whole network, even surpassing some domain generalization algorithms across a number of OOD generalization tasks. This implies that the better utilization of pretrained weights plays a leading role in enhancing the evaluated performance, instead of bolstering the OOD generalization ability from training data to test data. 

At present, it remains a fundamental open problem that how we should design a more reasonable evaluation procedure in the presence of strong pretrained models. 
Yu et al. \cite{yu2023rethinking} modify the DomainBed protocol \cite{gulrajani2020search} by removing the pretrained weights and training all models from scratch when evaluating OOD generalization algorithms. This serves as a pure and accurate evaluation protocol, but increases the evaluation burdens due to training large models from scratch. 
Another dataset ObjectNet \cite{barbu2019objectnet}, though not specially designed for OOD generalization, collects the data by taking new photos with various attributes including rotation, viewpoint, and background. In this way, they ensure that the constructed dataset has not appeared in any model's training data before. ImageNet-pretrained models suffer a huge performance drop on ObjectNet, confirming its success of preventing data leakage. However, datasets with images similar to ObjectNet may arise, and images in ObjectNet may be crawled and collected into newer datasets used for pretraining in the future, so the problem still exists by that time.

\section{Conclusion and Future Directions}
\label{sec:conclusion}

A scientific and reasonable paradigm of OOD evaluation, i.e. the evaluation of Out-of-Distribution Generalization, is of paramount significance not only to promoting the development of new OOD generalization algorithms, but also to the more practical and flexible application of existing models. In this paper, we undertake a systematic review of the problem setting, three primary categories of OOD evaluation methods, and briefly discuss in the context of pretrained models. 
Although we have discussed drawbacks and future directions for each specific category of OOD evaluation, here we enumerate several valuable potential future directions that are more general but somewhat absent in the current literature, based on our perception and reflection of OOD evaluation at present. 

\paragraph{OOD evaluation beyond performance}
Most existing OOD evaluation paradigms predominantly center around the performance of models. They aim to evaluate whether their OOD performance is good enough or not, or make comparisons among models. Nonetheless, as stated in Section \ref{sec:intro}, the objective of OOD evaluation is more than "whether", but also "where". Since the ultimate solution of OOD generalization is hard to find considering various kinds of distribution shifts, it is more practical in the deployment of machine learning models to identify the safe or risky regions for a specific model and to certify the kind of distribution shift that an algorithm can effectively address. A few works mentioned above \cite{thams2022evaluating, zhang2022did, kulinski2023towards, cai2023diagnosing, liu2023need} serve as primitive efforts towards this objective. It is worth noting that they are all relatively new works, suggesting that there may be an ongoing trend towards the evaluation beyond performance.

\paragraph{OOD evaluation of training data}
While fully-trained models are a composite result of model structures, algorithms, and training data, current OOD evaluation paradigms mostly evaluate models across different model structures or different algorithms, but seldom across different kinds of training data, and they seldom analyze the properties of training data that is beneficial to OOD generalization. 
In an era when people pay attention to developing new model structures and new algorithms, training data is easy to ignore, despite its pivotal impact on model performance \cite{mazumder2022dataperf,zhang2022nico,zha2023data}. Only a limited number of studies delve into this issue from the perspective of training data heterogeneity \cite{liu2022measure,shen2023meta}. 

\paragraph{Distinguishing performance of OOD generalization from ID generalization}
Existing OOD evaluation methods rely on direct comparisons of absolute performance. However, we raise concerns about whether these comparisons accurately gauge the authentic OOD generalization capability of models. There is a chance that the performance gain is attributed to the improvement of ID generalization ability, instead of OOD generalization ability. 
For example, with sufficient training on the same dataset, a larger network tends to achieve a higher performance on both ID and OOD test data. The performance gap between ID and OOD, though not necessarily the proper evaluation metric, may remain the same or even get larger. 
Of course, increasing the size of training data and the model capacity usually helps increase ID performance, thus increasing OOD performance, but it does not mean this is the answer to solving the OOD generalization problem, considering that large models also suffer from severe performance degradation encountering distribution shifts \cite{yang2022glue,yuan2023revisiting}, as well as the related issues of bias and fairness \cite{zhao2023survey,chang2023survey}. 
Therefore, for the ultimate solution of OOD generalization, we suggest that we should distinguish the OOD performance from ID performance when evaluating the OOD generalization ability of models. Simply using the performance gap might not be suitable since strong regularization tricks like using a large weight decay may decrease the ID performance to reduce the performance gap. There are already several works discussing the relationship between OOD and ID performance \cite{taori2019robustness,mania2020classifier,miller2021accuracy,andreassen2022evolution,yuan2023revisiting} while deeper investigations concerning this aspect are expected.

\ifCLASSOPTIONcaptionsoff
  \newpage
\fi

\bibliographystyle{IEEEtran}
\bibliography{IEEEabrv,tkde}

\end{document}